\documentclass{bmvc2k}

\usepackage{soul}
\usepackage{url}
\usepackage[small]{caption}
\usepackage{graphicx}
\usepackage{subcaption} 
\usepackage{amsmath}
\usepackage{amsthm}
\usepackage{multirow}
\usepackage{makecell}
\usepackage{booktabs}
\usepackage{algorithm}
\usepackage{algorithmic}
\usepackage{color}
\usepackage{indentfirst}

\usepackage{wrapfig}

\newcommand{\eg}{e.g.}
\newcommand{\ie}{i.e.}
\AtBeginDocument{\hypersetup{pdfborder={0 0 0}}}

\title{EAPruning: Evolutionary Pruning for Vision Transformers and CNNs}

\addauthor{Qingyuan Li}{liqingyuan02@meituan.com}{1}
\addauthor{Bo Zhang}{zhangbo97@meituan.com}{1}
\addauthor{Xiangxiang Chu}{chuxiangxiang@meituan.com}{1}

\addinstitution{
Meituan \\
Beijing, China
}
\runninghead{Li et al.}{EA-Pruning}

\def\eg{\emph{e.g}\bmvaOneDot}

\begin{document}

\maketitle

\begin{abstract}

Structured pruning greatly eases the deployment of large neural networks in resource-constrained environments. However, current methods either involve strong domain expertise, require extra hyperparameter tuning, or are restricted only to a specific type of networks, which prevents pervasive industrial applications.  In this paper, we undertake a simple and effective approach that can be easily applied to \emph{both vision transformers and convolutional neural networks}. Specifically, we consider pruning as an \emph{evolution process of sub-network structures that inherit weights through reconstruction techniques}. We achieve 50\% FLOPS reduction for ResNet50 and MobileNetV1, leading to \textbf{1.37$\times$} and \textbf{1.34$\times$} speedup respectively. For DeiT-Base, we reach nearly \textbf{40\%} FLOPs reduction and \textbf{1.4$\times$} speedup. Our code will be made available.

\end{abstract}

\section{Introduction}
\label{sec:intro}
With the large-scale application of AI in the cloud, mobile, autonomous vehicles, and IoT devices, deep neural networks face the challenge of reducing the high computational intensity and great power consumption on restricted devices. Researchers from both academia and industry  have endeavored on model compression and acceleration. A large volume of acceleration methods has been proposed. Some design lightweight networks like manually like MobileNet \cite{howard2017mobilenets} and ShuffleNet \cite{zhang2018shufflenet}, or through \emph{neural architecture search} (NAS) \cite{howard2019searching}, which requires deep expertise to invent new structures or to design appropriate search spaces. In contrast, pruning from pretrained networks can be put into direct use. To name a few, one can resort channel pruning \cite{he2017channel}, network trimming \cite{hu2016network}, data-free $l_1$-norm filter pruning \cite{li2016pruning}, structured sparsity learning \cite{wen2016learning}, and greedy search-based ThiNet \cite{luo2017thinet} etc.


In this paper, we attempt to further investigate pruning for its versatility and efficiency. Due to the large number of neural network parameters, the existence of redundant neurons becomes inevitable. It has been a natural idea to cut off ``unimportant'' neurons. Methods fall in this category can be roughly summarized as \emph{heuristic-based} and \emph{learning-based}. The former includes $l_1, l_2$ regularization \cite{han2015learning,li2016pruning}, deep compression \cite{han2015deep} etc. The latter comprises network slimming \cite{liu2017learning} and fine-grained sparse pruning \cite{zhou2021learning}. However, according to AlexNet \cite{krizhevsky2012imagenet}, VGG \cite{simonyan2014very}, ResNet \cite{he2016deep}, it is agreed that network structure plays a decisive role in performance than neurons' position. This observation is echoed by \cite{liu2018rethinking}. With the emergence of automated exploration methods like AMC \cite{he2018amc}, NetAdapt \cite{yang2018netadapt}, MetaPruning \cite{liu2019metapruning}, and APQ \cite{wang2020apq}, ``structure'' search has gradually become the mainstream.

Therefore, we also consider pruning as a process of seeking extreme values in a finite space. Despite above-mentioned advances, searching-based approaches have been hindered for long, since the searching space is limited but still very huge. It becomes impractical to exhaustively evaluate every structure, which goes exponential  ($2^{N}$, given N as the channel number). However, the optimal choices of ``solutions'' (\ie{} network structures)  tend to be similar, for which we can resort to evolutionary algorithm (EA) for searching. EA keeps good structures and discards bad ones until we gradually approach the optimal solution. Specifically,  we encode channel by its number as in \cite{liu2019metapruning} instead of channel-wise encoding \cite{wang2017towards,fernandes2021pruning,zhou2019knee} to achieve  compression of evolutionary space,  we then use weight reconstruction \cite{he2017channel} instead of ``cumbersome'' fine-tuning to recover subnetworks for proxy evaluation. We finally employ a multi-objective evolutionary process NSGA-III \cite{deb2013evolutionary} to better balance the computation and accuracy trade-off.


In a nutshell, our contribution can be summarized as follows,

\begin{itemize}
\item We propose an evolutionary-based pruning method (called EAPruning) for both vision transformers and CNNs, which is proved very effective, easy to use, and comes with low cost.  To our best knowledge, it is also the first pruning algorithm that \emph{works for both types of mainstream networks}.
\item Our method employs \emph{weight reconstruction} for the fast evaluation of sampled child networks, and \emph{multi-objective evolutionary search} to select a series of target models at once. Benefitting from this paradigm, we don't require tedious hyperparameter tuning and much domain expertise.
\item We prune DeiT-Base by nearly \textbf{40\%} FLOPs reduction with only 0.5\% loss in accuracy, and ResNet50 by \textbf{50\%} FLOPs reduction with merely 0.3\% accuracy loss.
\end{itemize}

\section{Related work}
\label{sec:rw}
Pruning is usually divided into \emph{unstructured pruning} \cite{han2015learning,han2015deep,guo2016dynamic,frankle2018lottery,dai2019nest,kusupati2020soft} and \emph{structured pruning} depending on the granularity of connections. The former prunes every connection to achieve sparse kernels in the end that won't enjoy much benefit unless it is efficiently supported by the underlying hardware. Therefore, most community effort is put into structured pruning.

\textbf{Criteria-based Pruning.} Structured pruning usually removes redundant filters according to some human-crafted criteria.  \cite{li2016pruning} directly identifies the relative importance of filters by their $l_1$ norms. Instead, \cite{hu2016network} calculates the percentage of activations of a neuron. Some use regularization terms as in \cite{wen2016learning} to sparsify the network where filters have smaller norms  are considered less important and can be safely pruned.   Other strategies like network slimming \cite{liu2017learning} penalize the scaling factor in batch normalization to find the unnecessary channels. In contrast, \cite{molchanov2019importance} involves Taylor expansion to determine the importance of filters. These methods commonly come with costly iterative finetuning and require predefined pruning ratios or empirics to obtain appropriate compression. ResRep \cite{ding2021resrep} is another form of criteria that divides neurons into remembering parts and forgetting ones, but it is limited to CNN only. 

\textbf{Search-based Pruning.} \cite{liu2018rethinking} suggests that it is useful to consider pruning as an architecture search paradigm. Automated pruning methods have also begun to emerge, such as reinforcement learning in AMC \cite{he2018amc}, greedy search in NetAdapt \cite{yang2018netadapt}, neural architecture search in OFA \cite{cai2019once}, APQ \cite{wang2020apq}, and BigNAS \cite{yu2020bignas}. Note OFA and BigNAS benefits from supernet training strategy so that subnetworks with inherited weights don't require finetuning any further.

%

\textbf{Vision Transformer Compression.}  Making vision transformers \cite{dosovitskiy2020image} more efficient is an urgent task. VTP \cite{zhu2021visual} adds soft gates to learn the importance of filters. NViT \cite{yang2021nvit} instead imposes Taylor importance score.  AutoFormer \cite{chen2021autoformer} applies one-shot weight-sharing neural architecture search to find optimal subnetworks in predefined discrete search spaces. 

\section{Method}\label{sec:method}

\subsection{Motivation}
A universal pruning framework is indispensable to satisfy diverse model compression needs. Given the success of vision transformers and CNNs, we are driven to develop an efficient and that is easily applicable to both kinds of networks. It has been shown that structure is more important than weights \cite{liu2018rethinking}. There is no obvious difference for channels in the same layer, which disables a family of algorithms that tend to identify the importance of filters, such as Lasso \cite{he2017channel}, KL divergence \cite{yu2021unified}, Taylor expansion \cite{molchanov2019importance}.  In this regard, we'd like to resort to searching algorithms to find a better structure. However, previous evolutionary-based pruning methods face several challenges, such as \emph{huge search space} that is too fine-grained down to channel-wise level, and \emph{computationally expensive finetuning} for the evaluation of each subnetwork. Searching becomes very costly and infeasible for increasingly big models and large datasets.  To solve these problems, we need to choose a coarse pruning space for CNNs and to identify the effective prunable parts in Vision Transformer that relieves the searching difficulty. To get rid of costly iterative finetuning, meanwhile being general, we wish to apply minimum weight reconstruction to recover pruned structures. We also want to utilize evolutionary algorithms to enable proficient searching, while considering the computation vs. accuracy trade-off.

\subsection{Pruning Space}

Defining a proper pruning space is critical. The common channel-wise one-hot encoding is too fine-grained. Although it is useful to differentiate important channels from others for the iterative pruning paradigm, this creates unnecessary difficulty for searching-based algorithms. Take ResNet50 as an example, there are a total of 24576 channels excluding the first layer, whose search space is as huge as $2^{24576}$, effectively disabling evolutionary search. Instead, we follow the practices in neural architecture search that encodes channel numbers only. Therefore, for ResNet50, we need $\sum \log_2 (C_i) \approx 286$ bit long encoding, where $C_i$ is the number of channels of the $i$-th filter.

\begin{figure}[ht]
	\centering
	\includegraphics[width=0.5\columnwidth]{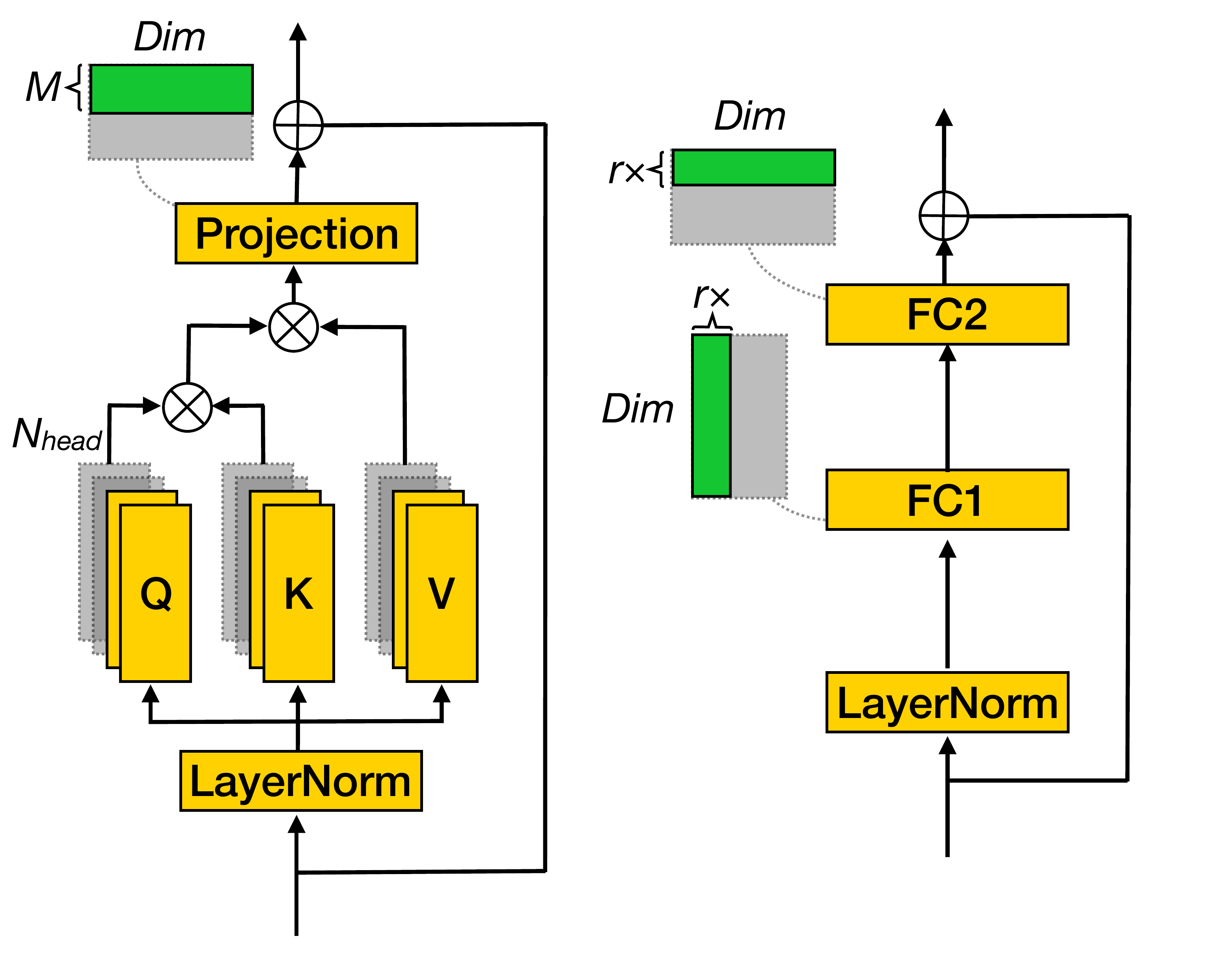}
	\vskip 0.1in
	\caption{Our pruning space for an attention block in Vision Transformers. We prune the number of heads for Q, K and V (correspondingly the projection dimension) and MLP inner dimension only. The dotted gray area shows weights to be pruned, green area shows the remaining weights for convolutions. The shape of input and output of each block is retained. }
	\label{fig:vit-pruning-space}
	\vskip -0.1in
\end{figure}

For vision transformers, we find it is enough to reduce the prunable space to the number of attention heads ($N_{head}$) and MLP ratios ($r$), see Figure~\ref{fig:vit-pruning-space}. We keep the embedding dimension ($Dim$) unchanged for large models like DeiT-Base as later shown these models tend to be ignored very early in evolution. Specifically for each attention block, we prune the number of heads for Q, K, and V, whose weights are pruned to $N_{head} \times Dim_{head} \times Dim$. Correspondingly, the weights in projection $Dim \times Dim$ are pruned to  $M \times Dim$, where $M=N_{head} \times Dim_{head}$. For MLP layers, we tune the hidden dimension by the ratio $r$ where the weights of FC1 are pruned to $Dim \times r \times Dim_{hidden}$, and FC2 to $ r \times Dim_{hidden} \times Dim$.

\subsection{Channel Selection}
Since we only encode the number of channels, it is still a problem to decide which channels to preserve. Common practices use $l_1$-norm as in AMC \cite{he2018amc}, lasso regression as in channel pruning \cite{he2017channel} , or greedy search as in NetAdapt \cite{yang2018netadapt} and ThiNet \cite{luo2017thinet}.  As we assume that the performance of the subnet only has to do with the structure, we just randomly sample channels to a target number. We later show that using $l_1$-norm doesn't make much difference. 

\subsection{Weight Reconstruction}
Direct channels pruning will cause performance collapse for subnetworks. To resolve such collapse, it is common to apply a finetuning process as in \cite{han2015learning,han2015deep,yang2018netadapt}. Despite being time-consuming especially on large datasets like ImageNet, the subnetwork after finetuning has not completely converged. 
To avoid the performance collapse and to have a justified evaluation of subnetworks, we use the technique of network \emph{weight reconstruction} as in \cite{he2017channel,luo2017thinet,he2018amc} to recover the remaining weights.

Specifically, given $F_{N \times C_{in} \times H \times W}$ is the input feature, $W_{C_{out} \times C_{in} \times K_1 \times K_2}$ is the convolutional kernels. To save computational cost, we randomly select $d$ patches $X_{N \times d \times C_{in} \times K_1 \times K_2}$ from $F$, whose corresponding output feature is $Y_{N \times d \times C_{out}} = X \times W^T$. We then sample $C_{in}'$ channels from $X$ to have $X'_{N \times d \times C_{in}' \times K_1 \times K_2}$, the weights of convolution kernel becomes $W'_{C_{out} \times C_{in}'  \times K_1 \times K_2}$, hence the output is $Y'_{N \times d \times C_{out}} =X' \times W'^T$. We seek to have $Y' = Y$ after channel selection. To do so, we have to update the weights $W$ that minimizes the reconstruction error as defined in  Equation~\ref{eq:linear-regression}.

\begin{equation}
	\begin{aligned}
		W' =\underset{W'}{\arg \min } \sum_{i=1}^{m}\left(Y-X' W'^{{T}} \right)^{2}
		\label{eq:linear-regression}
	\end{aligned}
\end{equation}

Such a linear regression problem can be solved by the common least-square method to find $W'=(X'^{T} X')^{-1}X'^{T}Y$. In addition, since the finetuning on the dataset is eliminated, the evolutionary  speed can be greatly improved.


\subsection{Pruning As Searching}
\begin{figure*}[ht]
\centering
\begin{subfigure}[b]{0.75\textwidth}
    \centering
    \includegraphics[width=\textwidth]{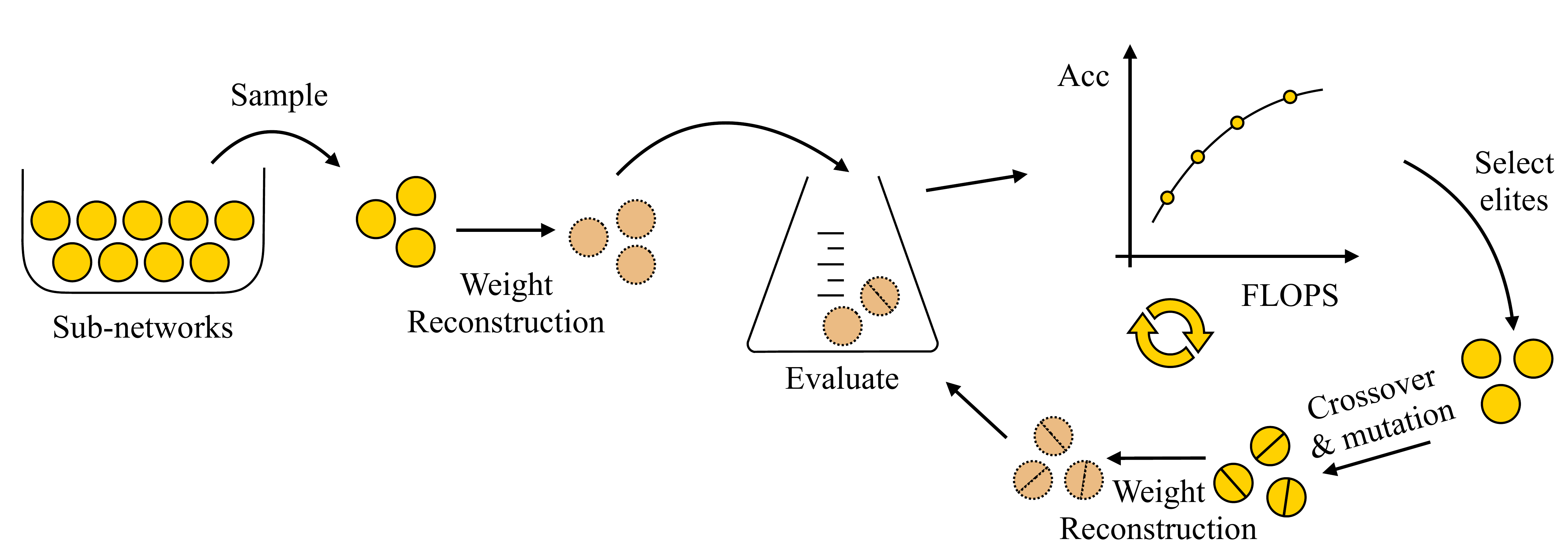}
\end{subfigure}
\begin{subfigure}[b]{0.22\textwidth}
	\centering
	\includegraphics[width=\textwidth]{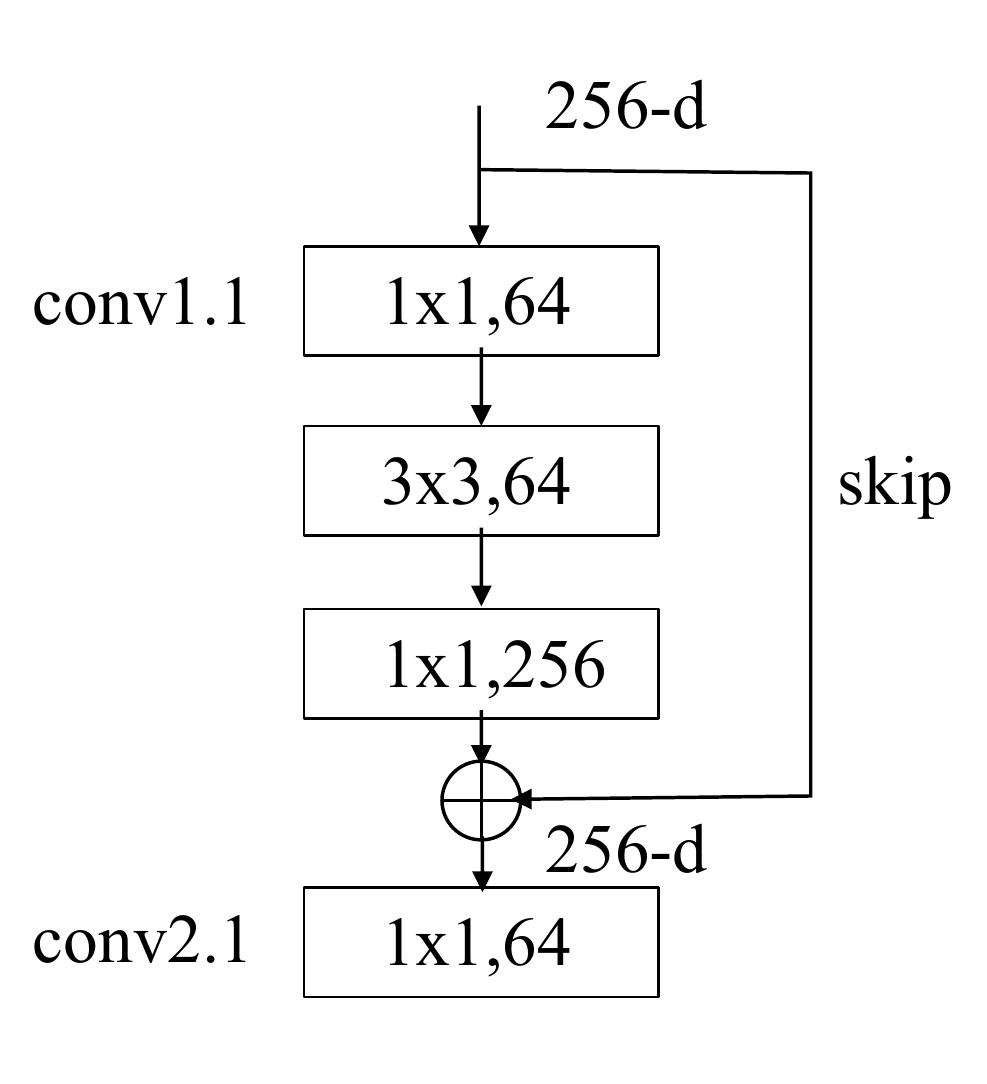}
\end{subfigure}
\vspace{0.1in}
	\caption{(a) Our evolutionary pruning pipeline. The reparameterization is adopted for fast evaluation within NSGA-III iterations. (b) Filters with channel dependency (\eg conv1.1 and conv2.1) shall be pruned correspondingly. }
	\label{fig:evolution-pipeline}
	\vskip -0.2in
\end{figure*}

We choose NSGA-III \cite{deb2013evolutionary} as the evolutionary algorithm, instead of vanilla evolution for it has two following advantages. Firstly, it introduces a reference point of the hyperplane to maintain the diversity of the population in the evolution process, which is very important to find a set of ``optimal network structures''. Secondly, after the evolutionary process ends, we can sample multiple subnets that meet  different constraints from the Pareto front. 
The knee-guided algorithm is also excluded \cite{zhou2019knee}, because its goal is to find a compromise in the ``multi-objective'', but we want to find different optimal solutions instead. Our whole pipeline can be viewed in Figure~\ref{fig:evolution-pipeline} and summarized in Algorithm~\ref{alg:evo-prune}.

\begin{algorithm}[ht]
\caption{Evolutionary Pruning Algorithm}
\label{alg:evo-prune}
\textbf{Input}: Original Network:  $\mathcal{N}$, Pretrained Weights: $\mathcal{W}$, Population Size: $\mathcal{P}$, Number of Mutation: $\mathcal{M}$, Number of Crossover: $\mathcal{S}$, Max Number of Iterations: $\mathcal{T}$ . \\
\textbf{Output}: $\mathcal{K}$ optimal Sub-Networks: $\mathcal{G}_\mathcal{K}$ . \\
\begin{algorithmic}[1]
\STATE $\mathcal{G}_0$ = Random($\mathcal{N}$, $\mathcal{P}$);
\FOR{$i = 1:\mathcal{T}$}
\STATE $\mathcal{G}_{metric}$ = Infer(Reconstruct($\mathcal{G}_{i-1}$, $\mathcal{W}$));
\STATE $\mathcal{G}_i$ = NSGA-III.NextGen($\mathcal{G}_{metric}$, $\mathcal{M}$, $\mathcal{S}$);
\ENDFOR
\STATE $\mathcal{G}_\mathcal{K}$ = ParetoFront($\mathcal{G}_{\mathcal{T}}$, $\mathcal{K}$);
\STATE \textbf{return} $\mathcal{G}_\mathcal{K}$;
\end{algorithmic}
\end{algorithm}

\section{Experiment}\label{sec:exp}
\subsection{Models and Datasets}
We conduct the experiments on a variety of networks, specifically, MobileNet-V1, ResNet50, and DeiT-Base. For pure CNN networks, we let every channel be prunable, \ie, for each layer we select from a range of $N$ channels with step 1. Note channel dependency is required, see Figure~\ref{fig:evolution-pipeline} (b). We directly perform evolutionary search on ImageNet, however only a small portion is used. We randomly select a set of images (1.5k for ResNet50 and DeiT-Base, 3k for MobileNet-V1) for weight reconstruction, and another 5k images for the evaluation of  subnetworks' performance. We take images from the training set to avoid overfitting. 

\subsection{Evolutionary configurations}

We evaluate nearly 1500 models per search space (takes 3.17 GPU hours on a
A100 machine). Table~\ref{table:ea-config} gives our configurations when performing evolutionary searching. We simply use a common set of hyper-parameters for three tasks (for ResNet50, it was due to an artifact of implementation, but the total number of models are $\approx$1.5k).

\begin{table}[ht]
\vskip -0.1in
\small
\begin{center}
\setlength{\tabcolsep}{3pt}
\begin{tabular}{ccccc}
\noalign{\smallskip}
\hline
\textbf{Search Space} & \textbf{Initial} & \textbf{Population} & \textbf{Iterations}  & \textbf{Total} \\
\hline
MobileNet-V1 & 64 & 50 & 30 & 1564 \\
ResNet50 & 64 & 32 & 47 & 1568 \\
DeiT-Base & 64 & 50 & 30 & 1564 \\
\hline
\end{tabular}
\end{center}
\caption{Evolutionary configurations of selected search spaces.}
\label{table:ea-config}
\vskip -0.3in
\end{table}

\subsection{Details for pruning DeiT-Base}
\textbf{Search space.} Vision Transformer ~\cite{dosovitskiy2020image} is stacked with multiple transformer encoders. Each encoder is composed of a multi-head self-attention (MHSA) module and linear projection layers. Each has the same head number and head dimension. 
We choose to prune the popular DeiT-Base, whose prunable parts are a total of 4 linear projection layers: $FC_{qkv}$, $FC_{proj}$, $FC_1$, and $FC_2$. We follow the similar encoding paradigm as done in CNNs. 

\textbf{Weight reconstruction.} 
Given input resolution is 224$\times$224 and patch size 16$\times$16, we have 197 tokens (including $cls$). To speed up reconstruction, we only choose 20 tokens which include $cls$ token, and another 19 randomly sampled ones for weight reconstruction. 


\textbf{Finetune configuration.} We use the same training hyper-parameters as DeiT-Base for finetuning, except that the number of epochs is reduced from 300 to 100, following VTP \cite{zhu2021visual}. Our results are shown in Table \ref{table:DeiT-B_FLOPs}. Though our performance is somewhat inferior compared with AutoFormer~\cite{chen2021autoformer}, EAPruning requires much less expertise and hyper-parameter tuning. Besides, AutoFormer is a NAS method for vision transformers only.

\begin{table}[ht]
\small
\begin{center}
\setlength{\tabcolsep}{1pt}
\begin{tabular}{ccccccc}
\noalign{\smallskip}
\hline
\textbf{Model} \ \ \ & \textbf{FLOPs} \ \ \ & \textbf{Reduction}  & \textbf{Top-1} \ \ \ & \textbf{Epochs} & \textbf{Training}  \\
\hline
DeiT-Base~(\cite{touvron2021training}) & 17.8G & - & 81.8\% & 300 & Scratch  \\
\hline
VTP~(\cite{zhu2021visual}) & 13.8G &22.4\% & 81.3\% & 100 & Finetune  \\
EAPruning (Ours)  & 13.5G & 24.2\% & 81.3\% & 100 & Finetune \\
\hline
AutoFormer~(\cite{chen2021autoformer}) & 11.0G& 38.2\% & 82.4\%  & 500  & Supernet  \\
EAPruning (Ours)  & 11.0G & 38.2\% & 81.6\% & 500  & Scratch \\
\hline
\end{tabular}
\end{center}
\caption{Pruning DeiT-Base on ImageNet, compared with state-of-the-art search-based methods.}
\label{table:DeiT-B_FLOPs}
\vskip -0.2in
\end{table}

\subsection{Details for pruning ResNet50}



\begin{figure}[ht]
    \centering
	\includegraphics[width=0.48\textwidth]{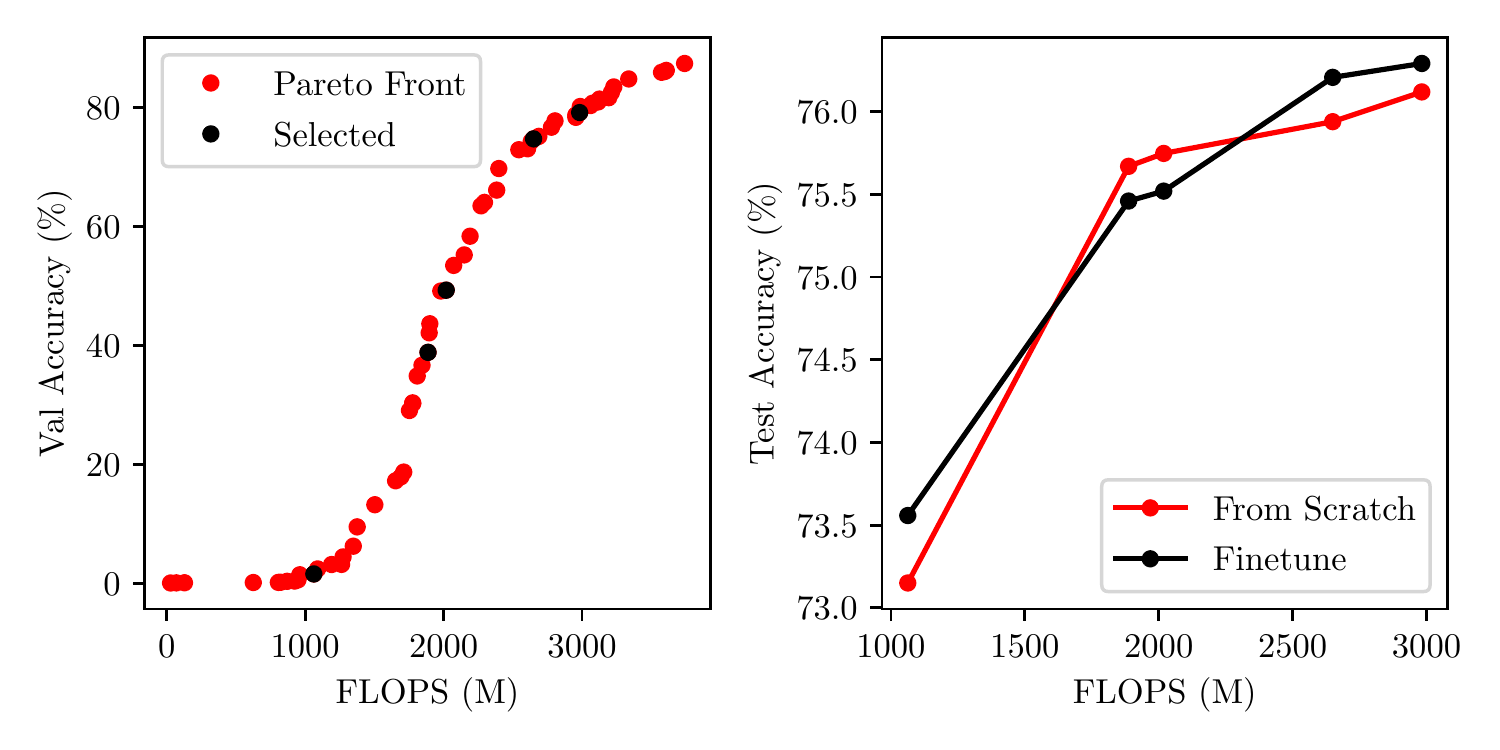}
	\vskip 0.1in
	\caption{(a) Pareto front when pruning ResNet50 with the proposed method. (b) Finetuning selected models (at different prunning ratios) vs. training them from scratch, both give the similar results.}
	\label{fig:r50-finetune-vs-scratch}
\end{figure}

\textbf{Training configuration.} According to \cite{liu2018rethinking}, we retrain all pruned models from scratch with the same recipe, an initial learning rate of 0.2 with cosine annealing for 200 epochs, a batch size of 512, and we weight decay to 1e-4. Table \ref{table:R50_FLOPs} shows our finetuned model compared with others. Note MetaPruning \cite{liu2019metapruning} and ResRep \cite{ding2021resrep}  are for CNNs only. Figure~\ref{fig:r50-finetune-vs-scratch} attests that finetuning and retraining give similar results (it is the structure that matters, not weight).

\begin{table}[ht]
\begin{center}
\small
\setlength{\tabcolsep}{1pt}
\begin{tabular}{cccc}
\noalign{\smallskip}
\hline
\textbf{Model} \ \ \ & \textbf{FLOPs} \ \ \  & \textbf{Reduction} & \textbf{Top-1}  \\
\hline
ResNet50~\cite{he2016deep} & 4111M & - & 76.0\% \\
\hline
AMC~\cite{he2018amc} & 2047M & 50.3\% & 75.5\% \\
NetAdapt~\cite{yang2018netadapt} & 2239M & 45.6\% & 75.9\% \\
MetaPruning~\cite{liu2019metapruning}  & 2G & 51.4\% & 75.4\%  \\
N2NSkip ~\cite{subramaniam2022n2nskip} & $\approx$2G  & 50\% & 74.6\%\\
ResRep~\cite{ding2021resrep} & $\approx$1871M$^{*}$ & 54.5\% &  76.2\% \\
EAPruning (Ours)  & 2019M & 50.9\% & 75.7\% \\
\hline
OTO~\cite{chen2021only} &  $\approx$1418M$^{*}$ & 65.5\% & 74.7\% \\
EAPruning (Ours)  & 1554M & 62.2\% & 74.8\% \\
EAPruning (Ours)  & 1063M & 74.1\% & 73.6\% \\
\hline
\end{tabular}
\end{center}
\caption{Pruned ResNet50 on ImageNet at 2G and 1G FLOPs level. $^*$: estimated by reduction ratio.}
\label{table:R50_FLOPs}
\end{table}

\subsection{Details for pruning MobileNetV1}


\begin{wraptable}{r}{0.5\textwidth}
\setlength{\tabcolsep}{1pt}
\small
\begin{center}
\begin{tabular}{cccc}
\noalign{\smallskip}
\hline
\textbf{Model} \ \ \ & \textbf{FLOPs} \ \ \ & \textbf{Reduction} & \textbf{Top-1} \\
\hline
MobileNetV1~(\cite{howard2017mobilenets}) & 569M &  & 70.6\% \\
\hline
MobileNetV1 ~(\cite{howard2017mobilenets}) & 325M & 0.75$\times$ & 68.4\% \\
AMC~(\cite{he2018amc}) & 301M & 0.89$\times$& 70.4\% \\
NetAdapt~(\cite{yang2018netadapt}) & 284M & 1$\times$& 69.1\% \\
MetaPruning~(\cite{liu2019metapruning})  & 281M & 1$\times$ & 70.6\%  \\
MetaPruning~(\cite{liu2019metapruning})  & 324M & 0.75$\times$ & 70.9\%  \\
EAPruning (Ours)  & 302M & 0.88$\times$ & \textbf{71.1\%} \\
\hline
\end{tabular}
\end{center}
\caption{Comparison of pruned MobileNetV1 models on ImageNet.}
\label{table:MV1_FLOPs}
\end{wraptable}

\textbf{Finetune configuration} We select a subnetwork of the same scale as AMC \cite{he2018amc} and finetune with the same strategy, \ie, an initial learning rate 0.05 with cosine annealing for 150 epochs, a batch size of 256, weight decay 4e-5. 

Table \ref{table:MV1_FLOPs} shows that our pruned network outperforms AMC by 0.6\% improved accuracy while having the same FLOPs. We also surpass MetaPruning \cite{liu2019metapruning} with fewer FLOPs.

\subsection{Ablation study}
\begin{figure}[ht]
\centering
\begin{subfigure}[b]{0.32\textwidth}
	\centering
	\includegraphics[width=\textwidth]{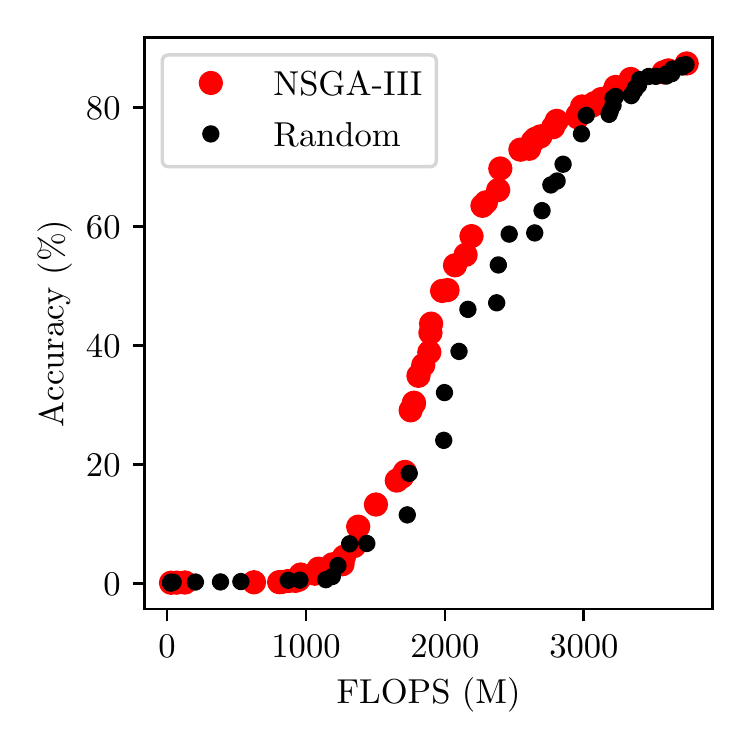}
\end{subfigure}
\begin{subfigure}[b]{0.32\textwidth}
	\centering
	\includegraphics[width=\textwidth]{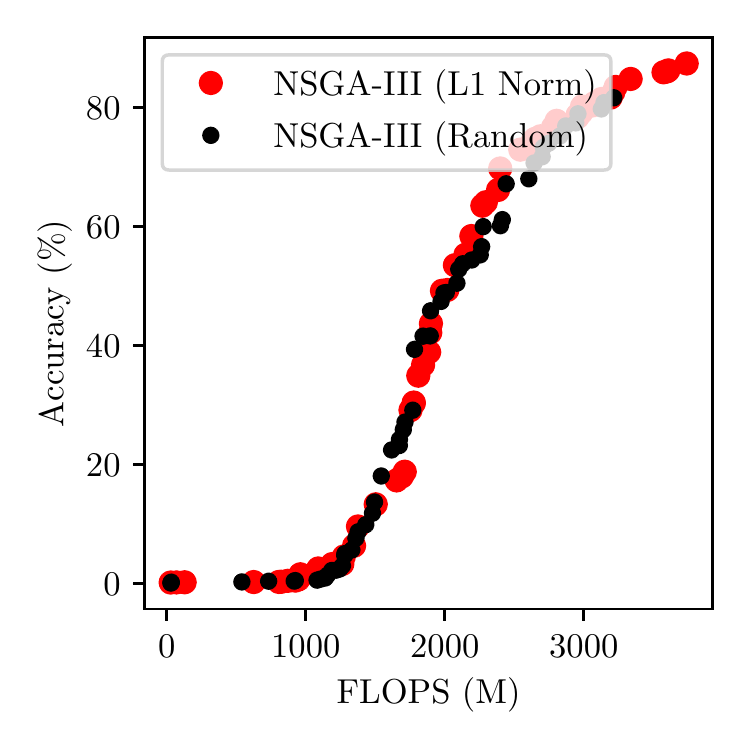}
\end{subfigure}
\vskip 0.1in
\caption{(a) Pareto front of NSGA-III vs. Random Searching (b) $l_1$-norm vs. random sampling.}
\label{fig:nsga-iii-vs-random}
\end{figure}

\textbf{NSGA-III vs. Random} To verify the effectiveness of the evolutionary algorithm, we randomly sample the same number of subnetworks from the ResNet50's search space, going through weight reconstruction and evaluation likewise, to obtain a new Pareto front. As shown in the Figure \ref{fig:nsga-iii-vs-random} (a), NSGA-III's advantage is obvious.

\textbf{$l_1$-norm vs. Random} To choose which channels to keep, we conduct a control experiment: $l_1$-norm vs. random sampling. As shown in Figure \ref{fig:nsga-iii-vs-random} (b), the Pareto fronts obtained by the two methods almost overlap, indicating that both can find subnetwork structures with the same performance. It is again assured that the network structure is more decisive for performance than weights.

\begin{wrapfigure}{r}{0.5\textwidth}
	\centering
	\includegraphics[width=0.42\columnwidth]{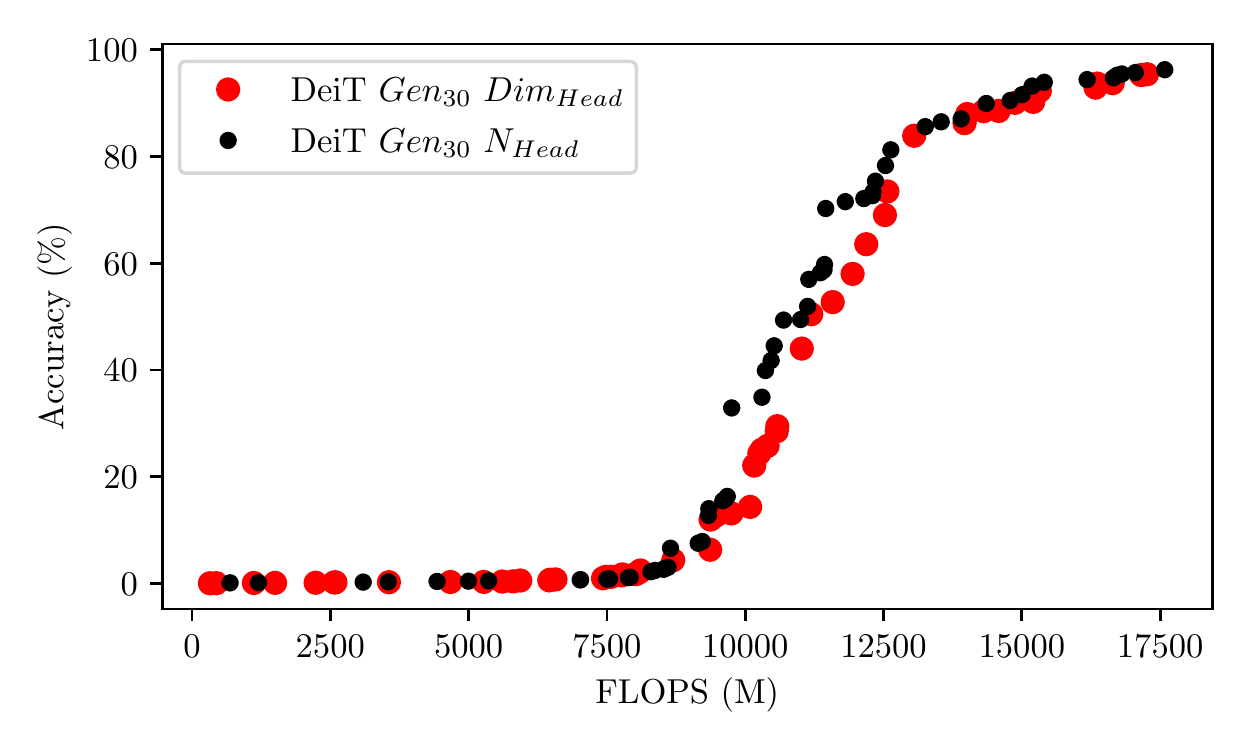}
	\vskip 0.1in
\caption{Pareto-front at the 30th generation when pruning DeiT-Base by head number and dimension.}
\label{fig:headnum-vs-headdim}
\end{wrapfigure}

\textbf{Head number vs. dimension} For DeiT-Base pruning, \cite{zhang2021rest} concludes that each head is only responsible for a projection subspace. To cut off the subspaces that have not learned effective information, by default we fixed the subspace size ($Dim_{Head}$) in the MSA of each block, and only pruned the number of projection spaces ($N_{Head}$). We also carried out a comparative experiment, fixing $N_{Head}$ in MSA and pruning $Dim_{Head}$ to reduce the size of the subspace. It can be seen from Figure~\ref{fig:headnum-vs-headdim} that pruning $N_{Head}$ is better than $Dim_{Head}$. The model of the same FLOPs is selected from the last generation Pareto boundary for finetuning, which also confirms the case, see Table~\ref{table:head_number_vs_dim}.  We can see that  shortening the subspace size of the heads in the same encoder will damage those heads that have learned effective information \cite{zhang2021rest}. 

\begin{table}[ht]
\setlength{\tabcolsep}{3pt}
\small
\begin{center}
\begin{tabular}{cccc}
\noalign{\smallskip}
\hline
\textbf{Type} \ \ \  & \textbf{FLOPS}  \ \ \ & \textbf{Top-1} \ \ \ & \textbf{Training}  \\
\hline
Head Num  & 12.5G   & \textbf{80.93\%}  & finetune  \\
Head Dim   & 12.5G   & 80.57\% & finetune \\
\hline
Head Num  & 10.3G   & \textbf{80.02\%} & finetune  \\
Head Dim   &  10.3G  & 79.74\% & finetune \\
\hline
Head Num  & 13.5G  & \textbf{81.45\%}  &  from scratch \\
Head Num   & 13.5G  & 81.28\% & finetune  \\
\hline
Head Num   & 10.3G  &  \textbf{81.27\%} & from scratch  \\
Head Num  & 10.3G &  80.02\% & finetune  \\
\hline
\end{tabular}
\end{center}
\caption{Pruning DeiT-Base by head number vs. dimension, either finetuned or trained from scratch.}
\label{table:head_number_vs_dim}
\vskip -0.1in
\end{table}

\textbf{Finetuning vs. Training from scratch}
To verify that structure of the network is more important than weights on Vision Transformers, we retrained the pruned DeiT-Base model from scratch using the same hyper-parameters as ~\cite{touvron2021training}. The finetuning uses the same recipe, except that the epoch is set to 100. The results are shown in Table \ref{table:head_number_vs_dim}. After training, the accuracy of the 13.5 GFLOPs model trained from scratch is improved by 0.17\% compared to finetuning, and the 10.3 GFLOPs model is improved by 1.25\%.

\begin{wraptable}{r}{0.5\textwidth}
\setlength{\tabcolsep}{1pt}
\small
\begin{center}
\begin{tabular}{cccc}
\noalign{\smallskip}
\hline
\textbf{Model} \ \ \ & \textbf{Throughputs}  \ \ \ & \textbf{Acc} \ \ \ & \textbf{Speedup} \ \ \ \\
& (img/s) & (\%) & \\
\hline
ResNet50 & 3753 & &  \\
ResNet50$\times$0.5& 5147  & -0.30 & \textbf{1.37$\times$} \\
\hline
MobileNetV1 & 9176 & & \\
MobileNetV1$\times$0.5  & 12296  & -0.09 & \textbf{1.34$\times$}  \\
\hline
DeiT-Base  & 777 &  &  \\
DeiT-Base$\times$0.6 & 1086 & -0.35 & \textbf{1.40$\times$}   \\
\hline
\end{tabular}
\end{center}
\caption{Our EA pruned models enjoys obvious speedup on NVIDIA A30 GPUs.}
\label{table:A30_QPS}
\end{wraptable}

\textbf{Hardware speedup}
The acceleration on hardware is an important factor for pruning. We conducted experiments on NVIDIA A30 GPU which has FP32 Tensor cores. We set model input to 128$\times$3$\times$224$\times$224, with FP32 precision, and we use TensorRT for deployment. The throughput of pruned CNN models has +30\% increase while the accuracy loss is less than 0.5\%. The throughput of DeiT is increased by 40\% at the pruning ratio  40\%, see Table \ref{table:A30_QPS}.

\section{Importance of the Projection Layers in DeiT}
\label{sec:dis}
\begin{figure*}[ht]
\vskip -0.2in
	\centering
	\includegraphics[width=\textwidth]{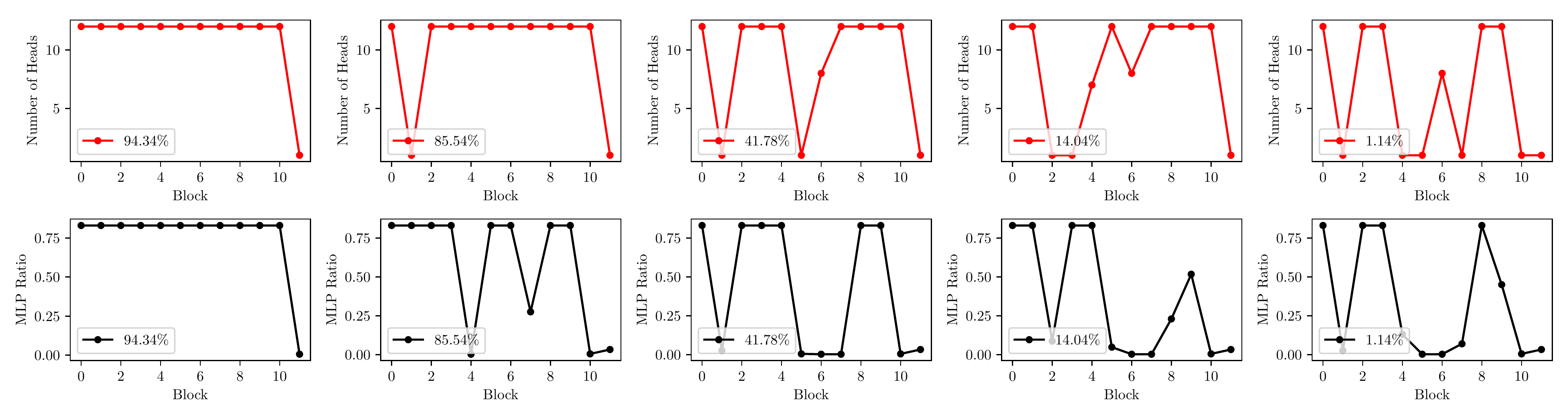}
\vskip 0.1in
\caption{Head numbers and MLP ratios of EA pruned DeiT-Base selected from the Pareto front. The legend shows the top-1 accuracy on a reserved small dataset after weight reconstruction.}
\label{fig:vit-preserve-per-block}
\vskip -0.05in
\end{figure*}

For DeiT-Base, we select networks with different pruning ratios from the Pareto front for structural analysis. Shown in Figure~\ref{fig:vit-preserve-per-block}, EA prunes the last layer first, and the layer near the head next, forming a ``cone'' structure~\cite{yang2021nvit}. As the pruning rate further increases, it starts to prune the projection layer in the middle of the network, forming a ``double-hump'' structure. Hence we speculate that the importance of the projection layer of DeiT for ImageNet classification can be ranked as follows: last layer $<$ beginning layers  $<$  middle layers $<$ other layers. This observation coincides with \cite{zhou2021deepvit}, which finds that deeper layers in vision transformers develop similar attention maps and the performance saturates thereafter.

\section{Conclusion}
\label{sec:con}
In this work, we have proposed a general pruning framework that is \emph{effective for both Transformers and CNNs}. We seek to rejuvenate the evolutionary search as an efficient and practical pruning paradigm by reducing the search space and utilizing weight reconstruction for fast evaluation of subnetworks. We also take advantage of the multi-objective nature of the NSGA-III. The entire pruning process is made simple enough, requiring less effort to tune hyper-parameters or to involve extra components. This is in line with the ``Occam's razor'' principle. Our method is evaluated on a variety of networks to achieve abundant reduction and speedup.

\bibliography{egbib}
\end{document}